\documentclass{article}
\newcommand{\bra}[1]{\bigl\langle #1 \bigr|}
\newcommand{\ket}[1]{\bigl| #1 \bigr\rangle}

\usepackage{epsfig}
\usepackage{graphics}
\usepackage[titletoc,toc,title]{appendix}
\renewcommand \appendix{\par
  \setcounter{section}{0}
  \setcounter{subsection}{0}
  \setcounter{figure}{0}
  \setcounter{table}{0}
   }

\begin{document}
\begin{center}
{\large Autonomous Quantum Perceptron Neural Network}

\vspace{0.5cm}
Alaa Sagheer \footnote{Corresponding author} and Mohammed Zidan \\
Department  of Mathematics\\
Center for Artificial Intelligence and RObotics (CAIRO)\\
Faculty of Science, Aswan University, Aswan,
Egypt\\
Email: \ alaa@cairo-aswu.edu.eg\\

\end{center}

{\bf Abstract}:Recently, with the rapid development of technology, there are a lot of applications require to achieve low-cost learning. However the computational power of classical artificial neural networks, they are not capable to provide low-cost learning. In contrast, quantum neural networks may be representing a good computational alternate to classical neural network approaches, based on the computational power of quantum bit (qubit) over the classical bit. In this paper we present a new computational approach to the quantum perceptron neural network can achieve learning in low-cost computation. The proposed approach has only one neuron can construct self-adaptive activation operators capable to accomplish the learning process in a limited number of iterations and, thereby, reduce the overall computational cost. The proposed approach is capable to construct its own set of activation operators to be applied widely in both quantum and classical applications to overcome the linearity limitation of classical perceptron. The computational power of the proposed approach is illustrated via solving variety of problems where promising and comparable results are given.
\section{Introduction}	        
\vspace*{-0.5pt}
\noindent
Classical Artificial Neural Networks (CANN) derives its computing power through its massively parallel-distributed structure and the ability to learn and, therefore, generalize. However, CANN may face many difficulties such as the absence of concrete algorithms and rules for specifying optimal design architectures, limited memory capacity, time-consuming training, etc.\cite{Hagan}. One of the known classical approaches is the classical Perceptron Neural Network (CPNN), which is applied only for linearly separable learning problems \cite{Rosenblatt}. In other words, CPNN cannot be applied for problems which have inseparable classes, such as XOR problem \cite{Hagan}.
These limitations, and others, have been motivated many researchers to investigate new trends in neural computation domain \cite{Shafee,Zhou}. One of the novel trends in this domain is to evoke properties and techniques of quantum computing into classical neural computation approaches.

Several researchers expect that quantum computing is capable to enhance the performance, and overcoming the above limitations, of classical neural computation \cite{Sagheer,Ventura,li,Nielsen}.
The beginning was in 1995 with Kak \cite{Kak} who was the first researcher introduced the concept of quantum neural computation. Then, Menneer \cite{Menneer} defines a class of quantum neural network (QNN) as a superposition of single component networks where each is trained using only one pattern. Ventura et al. in \cite{Ventura} introduced a new associative memory technique based on Grover's quantum search algorithm can solve the completion problem. The technique restores the full pattern when a part of it is initially presented with just a part of the pattern. Also, an exponential increase in the capacity of the memory is performed when it compared with the CANN capacity.

As long as classical perceptron is concerned, some researchers have been tried to increase the efficiency of perceptron using the power of quantum computation. In 2001, Altaisky \cite{Altaisky} developed a simple quantum perceptron that depends on selecting the activation operator. However its simplicity, Altaisky approach consumed much time in order to select an activation operator, especially, when the size of training data is large.
Next, Fei et al. \cite{Fei} introduced a new model of quantum neuron and its learning algorithm based on Altaisky perceptron. Fei model used the delta rule as the learning rule which yields considerable results such as computing XOR-Function using only one neuron and nonlinear mapping property. Unfortunately, Fei model did not provide us a new way for deriving the activation operator. Nevertheless, Fei model is sensitive for the selection of the appropriate activation operator, which was the problem of Altaisky perceptron.

Recently, Zhou et al. \cite{ZhouQ} developed a quantum perceptron approach based on the quantum phase adequately and could to compute the XOR function using only one neuron.The  drawback of Zhou perceptron is that it requires many computation iterations to give a response. Finally, Siomau \cite{Siomau} introduced an autonomous quantum perceptron based on calculating a set of positive valued operators and valued measurements (POVM). However, Simomau perceptron cannot be applied for problems such as quantum-Not gate and Hadamard gate.

In this paper, we propose a novel autonomous quantum perceptron neural network (AQPNN) approach can be used to solve both classical applications and quantum applications. The proposed AQPNN improves the computational cost of Altaisky quantum perceptron as well as the computational cost of Zhou quantum perceptron and its ability to learn the problems that Siomau quantum perceptron can not learn. The proposed perceptron is capable to
adapt its activation operator very fast which reduces the overall learning time. In addition, it is capable to overcome the linearity restriction of classical perceptron where AQPNN can be viewed as a non-linear perceptron. To evaluate AQPNN, we solve various problems and the results are compared favorably with Zhou \cite{ZhouQ}.

The paper is organized as follows: Section 2 describes the operation of AQPNN and its learning algorithm. Section 3 shows the computational power of AQPNN via solving various problems. Section 4 discusses the performanceof AQPNN. Section 5 shows the conclusion and our future work.

\section{Autonomous Quantum Perceptron Neural Network (AQPNN)}
\subsection{Description of the AQPNN}
\noindent
The proposed (AQPNN)approach is a quantum neural network approach includes only one neuron with \emph{n} qubit inputs $\ket{x_1}$, $\ket{x_2}$, $.....$, $\ket{x_n}$,(for qubit definition, see Appendix A). A set of weight operators ${w_1}$,${w_2}$, $.....$,${w_n}$ is assumed such that one weight operator is associated with each input, and $\ket{y_{net}}$ is the final network response; see Figure 1. The operators $F_j$ refers to a set of unique activation operators of the proposed perceptron.

\begin{figure}[h!]
  \begin{center}
  \includegraphics[width=20pc]{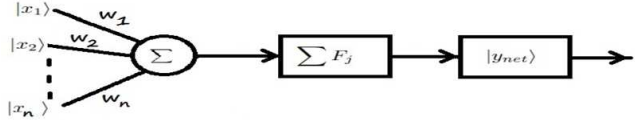}

           \caption{ The proposed AQPNN model}
       \end{center}
\end{figure}


The proposed AQPNN approach is based on a supervised learning procedure, that is, it is provided with a set of learning patterns (inputs/targets) in qubit form.

For each input pattern presented to the network, the weighted sum qubit $\ket{y_{j}}$ is calculated using the form:

\noindent
\begin{equation}
\ket{y_{j}}=\sum\limits_{i=1}^{n} {w_i} \ket{x_i}=\left[\begin{array}{*{1}{c}}
\alpha_j\\
\beta_j
\end{array}\right]
\,. \label{weighted sum}
\end{equation}
where $\alpha_j$ and $\beta_j$ are the probability amplitudes of the weighted sum qubit for $j^{th}$ pattern in the training set. The weight operators is updated at time \emph{t} using the following rule:
 \begin{equation}
w_i(t+1)=w_i(t)+\gamma \ket{e} \bra{x_i}
\,. \label{w-updated}
\end{equation}
where $\gamma$ is the learning rate, $\ket{e}=(\ket{d}-\ket{y})$ is the perceptron error and $\ket{e} \bra{x_i}$ denotes the outer product of vectors $\ket{e}$ and $\bra{x_i}$. Once the weighted sum is calculated for all the available patterns, then the set of activation operators can be calculated using the form:

 \begin{equation}
F_j=\left[\begin{array}{*{2}{c}}
\cos {\theta_j}& -\sin{\theta_j}\\
\sin {\phi_j}& \cos{\phi_j}
\end{array}\right]
\,. \label{activation}
\end{equation}
where $j = 1, 2, 3,...,m$, and $m$ is the set of unique activation operators (repeated activation operators are discarded) where $m\le N$ is the number of training data set. The parameters $\theta_j$ and $\varphi_j$ are two real valued angles calculated using the form:
\begin{equation}
\left[\begin{array}{*{2}{c}}
\cos {\theta_j}& -\sin{\theta_j}\\
\sin {\phi_j}& \cos{\phi_j}
\end{array}\right]\left[\begin{array}{*{1}{c}}
\alpha_j\\
\beta_j\end{array}\right]=\left[\begin{array}{*{1}{c}}
\alpha_{d_j}\\
\beta_{d_j}\end{array}\right]
\,. \label{angles}
\end{equation}
where [$\alpha_j$,$\beta_j]^{T}$ is the weighted sum qubit and ($\alpha_{d_j}$,$\beta_{d_j})^{T}$ is the target qubit.
The aim of each activation operator is to transform the weighted sum qubit to be mapped into the given target and make it a normalized qubit Eq.(\ref{quprob}). After calculating the set of all activation operators, the output of the autonomous quantum perceptron is given using the superposition of all activation operators in the following form:
\begin{equation}
\ket{y_{output}}=\sum\limits_{j=1}^{m}F_j \sum\limits_{i=1}^{n}\ket{x_i}
\,. \label{superposition}
\end{equation}
where $\ket{y_{output}}$ is the network output as a superposition of the set of output qubits. This output represents the effect of the activation operators (interference) on the weighted sum qubit resulted when any pattern presented to the network. One qubit only from these qubits will be the response of the network and can be specified by the following form:
\begin{equation}
\ket{y_{net}}=L(\left|(\bra{y_{output}}\circ \ket{y_{output}}-C)\right|)\circ \ket{D})
\,. \label{qubit}
\end{equation}
where $C=[1,1,...,1]^{T}$ is a good vector and $D$ is the vector of the target qubits. The operation $\circ$ achieves the Hadamard product operation \cite{Roger} (for more details, see Appendix A), where the function \emph{L} retain the smallest absolute value and makes it equal one and the rest of values to be equal zeros. Hence, the result of Eq.(\ref{qubit}) is only one qubit represents the net response of the AQPNN for the current input.

\subsection{The learning algorithm of AQPNN}
\noindent
According to the description given above, the AQPNN learning algorithm is divided into two main stages: The First stage is imbedded in both Eq.(\ref{angles}) and Eq.(\ref{superposition}), where the AQPNN algorithm collects information about the problem in hand by constructing a set of activation operators. In the second stage the AQPNN takes the decision about the network’s response according to Eq.(\ref{qubit}) based on the gathered information. In the following, we can summarize the AQPNN learning algorithm in the following steps:\\\\
\textbf{Step 1}: Set all $F_i = I$ (identity matrix). Then choose the initial weight operators $w_i$
randomly, set the learning rate $0<\gamma<1$ and set iteration number $k= 1$,\\\\
\textbf{Step 2}: Calculate the weighted sum qubit for each given pattern using Eq.(\ref{weighted sum}),\\\\
\textbf{Step 3}: Compare each weighted sum qubit for the patterns of each class with all other
weighted sum qubits for other classes. We have two cases here:
\noindent
\begin{enumerate}
  \item If each weighted sum qubit for any class does not equal the same value for any
weighted sum in any other classes then go to step 5,
else, go to step 4.
  \item If the value of any weighted sum qubit is zero then go to step 4,
else, go to step 5.
\end{enumerate}
\indent
\\
\textbf{Step 4}: Update the weight operators using Eq.(\ref{w-updated}),
set $k= k+1$, go to step 2. \\\\
\textbf{Step 5}: Calculate the activation operator for each weighted sum qubit using Eq.(\ref{angles}).

The superposition of output qubits of the network is given by Eq.(\ref{superposition}) whereas the net response of the AQPNN is given by Eq.(\ref{qubit}).

\section{The Computational Power of AQPNN}
\noindent
We proceed now to evaluate practically the computational power of the proposed AQPNN algorithm. In this section, we show the results of using AQPNN in solving four different problems. In the first two problems, we solve the problems of quantum Not-gate and the Hadamard-gate. In the third problem we compute the XOR-function, whereas in the fourth problem we achieve a classification task application.

\subsection{The Quantum Not-gate}
\noindent
\textbf{Class A}: The first pattern is $P_1 = \{ \ket{x_1}= \left[\begin{array}{*{1}{c}}
1\\
0\end{array}\right], \ket{d_1}= \left[\begin{array}{*{1}{c}}
0\\
1\end{array}\right]\}$

\noindent
\textbf{Class B}: The second pattern is $P_2 = \{ \ket{x_2}= \left[\begin{array}{*{1}{c}}
0\\
1\end{array}\right], \ket{d_2}= \left[\begin{array}{*{1}{c}}
1\\
0\end{array}\right]\}$

The initial weight operators is chosen arbitrary as $w=\left[\begin{array}{*{2}{c}}
1&0\\
0&1\end{array}\right]$.
Once, we introduced the two patterns to the AQPNN network we obtain the weighted sum, according to Eq.(\ref{weighted sum}), as follows:

\begin{center}
$\ket{y_1}= \left[\begin{array}{*{1}{c}}
1\\
0\end{array}\right], \ket{y_2}= \left[\begin{array}{*{1}{c}}
0\\
1\end{array}\right]$
\end{center}

Since $\ket{y_1}$ has a different value than $\ket{y_2}$, then we can calculate the set of activation operators as follows:\\
\begin{center}
$\left[\begin{array}{*{2}{c}}
\cos {\theta_1}& -\sin{\theta_1}\\
\sin {\phi_1}& \cos{\phi_1}
\end{array}\right]\left[\begin{array}{*{1}{c}}
1\\
0\end{array}\right]= \left[\begin{array}{*{1}{c}}
0\\
1\end{array}\right]$
\end{center}
\noindent
\\
i.e. $\theta_1=-90$, $\varphi_1= 90$ and, thereby, $F_1=\left[\begin{array}{*{2}{c}}
0&1\\
1&0\end{array}\right]$.
\indent
\\
Similarly, we can get $F_2$ as follows:\\
\begin{center}
$\left[\begin{array}{*{2}{c}}
\cos {\theta_2}& -\sin{\theta_2}\\
\sin {\phi_2}& \cos{\phi_2}
\end{array}\right]\left[\begin{array}{*{1}{c}}
0\\
1\end{array}\right]= \left[\begin{array}{*{1}{c}}
1\\
0\end{array}\right]$
\end{center}
\noindent
\\
i.e. $\theta_2=-90$, $\varphi_2= 90$ and, thereby, $F_2=F_1=\left[\begin{array}{*{2}{c}}
0&1\\
1&0\end{array}\right]$.\\
Therefore, the superposition output is $\ket{y_{output}}=F \sum\limits_{i=1}^{n} w_i \ket{x_i}$. This means that, the quantum– Not gate is trained after only one iteration.

\subsection{The Hadamard-gate}
\noindent
\textbf{Class A}: The first pattern is $P_1 =\{ \ket{x_1}= \left[\begin{array}{*{1}{c}}
1\\
0\end{array}\right], \ket{d_1}= \frac{1}{\sqrt2}\left[\begin{array}{*{1}{c}}
1\\
1\end{array}\right]\}$

\noindent
\textbf{Class B}: The second pattern is $P_2 = \{ \ket{x_2}= \frac{1}{\sqrt2}\left[\begin{array}{*{1}{c}}
0\\
1\end{array}\right], \ket{d_2}= \left[\begin{array}{*{1}{c}}
1\\
-1\end{array}\right]\}$

The initial weight operators is chosen arbitrary as $w=\left[\begin{array}{*{2}{c}}
1&0\\
0&1\end{array}\right]$
Once, we introduced the two patterns to the AQPNN network we obtain the weighted sum, according to Eq.(\ref{weighted sum}), as follows:

\begin{center}
$\ket{y_1}= \left[\begin{array}{*{1}{c}}
1\\
0\end{array}\right], \ket{y_2}= \left[\begin{array}{*{1}{c}}
0\\
1\end{array}\right]$
\end{center}

By the same way, as $\ket{y_1}$ has a different value than $\ket{y_2}$ then we can calculate the set of activation operators,we can get F1 as follows:\\
\begin{center}
$\left[\begin{array}{*{2}{c}}
\cos {\theta_1}& -\sin{\theta_1}\\
\sin {\phi_1}& \cos{\phi_1}
\end{array}\right]\left[\begin{array}{*{1}{c}}
1\\
0\end{array}\right]= \frac{1}{\sqrt2}\left[\begin{array}{*{1}{c}}
1\\
1\end{array}\right]$
\end{center}
\noindent
\\
i.e. $\theta_1=-45$, $\varphi_1= 135$ and, thereby, $F_1= \frac{1}{\sqrt2}\left[\begin{array}{*{2}{c}}
1&1\\
1&-1\end{array}\right]$.\\

Similarly, we can get F2 as follows:\\
\begin{center}
$\left[\begin{array}{*{2}{c}}
\cos {\theta_2}& -\sin{\theta_2}\\
\sin {\theta_2}& \cos{\theta_2}
\end{array}\right]\left[\begin{array}{*{1}{c}}
0\\
1\end{array}\right]= \frac{1}{\sqrt2}\left[\begin{array}{*{1}{c}}
1\\
-1\end{array}\right]$
\end{center}
\noindent
\\
i.e. $\theta_2=-45$, $\varphi_2= 135$ and, thereby, $F_2= \frac{1}{\sqrt2}\left[\begin{array}{*{2}{c}}
1&1\\
1&-1\end{array}\right]$.

Therefore, the superposition output is $\ket{y_{output}}=F \sum\limits_{i=1}^{n} w_i \ket{x_i}$. This means that, the Hadamard-gate is trained after only one iteration.

\subsection{The XOR-Function}
\noindent
It is known that, there are two classes in XOR-function:

\noindent
\textbf{Class A}: The first pattern is
 $P_1 =\{ \ket{x_1}=  \left[\begin{array}{*{1}{c}}
1\\
0\end{array}\right], \ket{x_2}=  \left[\begin{array}{*{1}{c}}
1\\
0\end{array}\right]\,\ket{d_1}=  \left[\begin{array}{*{1}{c}}
1\\
0\end{array}\right]\}$

\noindent
The second pattern is
$P_2 =\{ \ket{x_1}=  \left[\begin{array}{*{1}{c}}
0\\
1\end{array}\right], \ket{x_2}=  \left[\begin{array}{*{1}{c}}
0\\
1\end{array}\right]\,\ket{d_2}=  \left[\begin{array}{*{1}{c}}
1\\
0\end{array}\right]\}$

\noindent
\textbf{Class B}:
The third pattern is
 $P_3 =\{ \ket{x_1}=  \left[\begin{array}{*{1}{c}}
1\\
0\end{array}\right], \ket{x_2}=  \left[\begin{array}{*{1}{c}}
0\\
1\end{array}\right]\,\ket{d_3}=  \left[\begin{array}{*{1}{c}}
0\\
1\end{array}\right]\}$

\noindent
The fourth pattern is
$P_4 =\{ \ket{x_1}=  \left[\begin{array}{*{1}{c}}
0\\
1\end{array}\right], \ket{x_2}=  \left[\begin{array}{*{1}{c}}
1\\
0\end{array}\right]\,\ket{d_4}=  \left[\begin{array}{*{1}{c}}
0\\
1\end{array}\right]\}$

Assume a random initial weight operator takes the value $w1=w2=\left[\begin{array}{*{2}{c}}
1.1&1.2\\
0&0\end{array}\right]$. If we introduced the four patterns into the AQPNN network we obtain,after only one iteration, the weighted sum of each pattern as follows:\\

\begin{center}
$\ket{y_1}= \left[\begin{array}{*{1}{c}}
2.2\\
0\end{array}\right], \ket{y_2}= \left[\begin{array}{*{1}{c}}
2.4\\
0\end{array}\right] , \ket{y_3}= \left[\begin{array}{*{1}{c}}
2.3\\
0\end{array}\right]  , \ket{y_4}= \left[\begin{array}{*{1}{c}}
2.3\\
0\end{array}\right]$
\end{center}

\indent
\\
It is easy to observe that the first couple of weighted sum qubits has different values than the other couple of weighted sum qubits (i.e. either of $\ket{y_1}$  or $\ket{y_2} )$  has a different value than $\ket{y_3}$  and $\ket{y_4}$. Also, we may observe that $\ket{y_3}$ = $\ket{y_4}$ . As these two patterns are in the same class B, this implies that there only three activation operators will take the following forms:\\

$F_1= \left[\begin{array}{*{2}{c}}
0.4545& -0.8907\\
0&1
\end{array}\right] , F_2=\left[\begin{array}{*{2}{c}}
0.4167& -0.9091\\
0&1
\end{array}\right]  , F_3=F_4=\left[\begin{array}{*{2}{c}}
0& -1\\
0.4348&0.9005
\end{array}\right]$
\indent
\\
Then, the superposition output can be calculated as:

\begin{center}
$\ket{y_{output}}=\sum\limits_{j=1}^{m=3}F_j \sum\limits_{i=1}^{n=2} w_i \ket{x_i}$
\end{center}

Table \ref{t1} shows a comparison between the proposed perceptron AQPNN, Zhouh perceptron \cite{ZhouQ} and the classical perceptron, the classical pereptron is not applicable in case of using one neuron. The proposed perceptron gives the final output after only one iteration whereas Zhouh perceptron gives the final result after 16 iterations \cite{ZhouQ}. Then it is clear that AQPNN reduces the computation steps to get the final results.

\begin{table}[htbp]
\caption {A comparison between the proposed AQPNN perceptron, Zhouh perceptron and the classical perceptron to solve the XOR-function} \label{t1}
\centering

\begin{tabular}{|c |c |c |c|}
\hline
Algorithm name  &AQPNN  &Zhouh Perceptron &Classical Perceptron(One neuron)\\
\hline
No.of iterations&1 &16 &Not applicable by one neuron\\
\hline
\end{tabular}
\end{table}

%


\subsection{ Two-overlapped classification problem}
\noindent
We proceed now to use the proposed AQPNN approach in a classification application. The application we use here is atypical two-overlapped classes classification problem, which can be regarded as a complex generalization of the XOR problem \cite{Xiao};
see Figure 2. It has two classes: the first is a oval-shape class has the target  $\ket{0}= \left[\begin{array}{*{1}{c}}
1\\
0\end{array}\right]$
with arbitrary input patterns given in Table \ref{t2}. The second class is square-shape class which has the target $\ket{1}= \left[\begin{array}{*{1}{c}}
0\\
1\end{array}\right]$ with arbitrary input patterns given in Table \ref{t3}.

\begin{figure}[h!]
  \begin{center}
  \includegraphics[width=20pc]{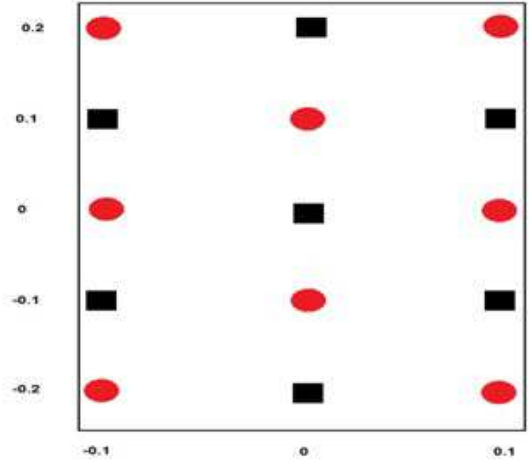}

           \caption{Two overlapped classes classification problem}
       \end{center}
\end{figure}

\begin{table}[htbp]
\caption {Training input patterns of the oval-shape class}
\label{t2}
\centering

\begin{tabular}{|c |c |c |c| c| c| c| c|}
\hline
P1  &P2  &P3   &P4  &P5  &P6  &P7  &P8  \\
\hline
(0.1,0)&(0.1,0.2) &(0,0.1) &(-0.1,0.2) &(-0.1,0) &(0,-0.1) &(0.1,-0.2) &(-0.1,-0.2)\\
\hline
\end{tabular}
\end{table}
\begin{table}[htbp]
\caption {Training input patterns of the square-shape class}
\label{t3}
\centering

\begin{tabular}{|c |c |c |c| c| c| c|}
\hline
P9  &P10  &P11  &P12  &P13  &P14  &P15     \\
\hline
(0.1,0.1)&(0,0)  &(0,0.2)  &(-0.1,0.1)  &(0.1,-0.1)  &(-0.1,-0.1)  &(0,-0.2)   \\
\hline
\end{tabular}
\end{table}

\noindent
It's clear that the values of input patterns are classical data, i.e. real values, so it must be transformed into qubits using qubit normalization equation Eq.(\ref{quprob}). For example, for the pattern P1= (0.1,0), where
a=0.1 and, then, $b=\sqrt{1-(0.1)^2}$ . Thus, we may have $P_1 =\{ \ket{x_1}= \left[\begin{array}{*{1}{c}}
0.1\\
0.9950\end{array}\right], \ket{d_1}= \left[\begin{array}{*{1}{c}}
1\\
0\end{array}\right]\}$. In this experiment, we chose only 15 patterns as training data whereas the testing data is generalized over 176 patterns. If the learning rate is chosen, randomly, to be 0.1, we will find the classification rate approaches $97.73\%$  after only one iteration for the learning process.



\section{Discussion}
\noindent
It is worth now to discuss the performance of the proposed algorithm. It is clear from the above examples that the computational power of the proposed AQPNN is high, however, many observations may be one will record. First observation is that under equal weight operators, the AQPNN model, in some applications, does not utilize all the training data like other perceptron algorithms \cite{Rosenblatt, ZhouQ, Siomau}. For example, in the first two situations, i.e. Not-gate and Hadamard-gate, we need only one training input (because we use only the unique activation operator), where as in the XOR function situation, it is required three training inputs in order to accomplish the learning process. In the three situations, the AQPNN is capable to reduce both the computation time and the number of activation operators.

Second observation is concerned with the relation between the initial weight operators and the activation operators. In the situations of the quantum Not-gate and the quantum Hadamard-gate, the initial weight operator was the unitary operator, whereas in case of the XOR-function it was not the unitary operator. The reason for this is due to the nature of the unitary operator\emph{U} where $UU^{\dagger}=I$. Then, using Eq.(\ref{activation}),  that includes the formula of the activation operators, we have:\\

\begin{center}
$\left[\begin{array}{*{2}{c}}
\cos {\theta_j}& -\sin{\theta_j}\\
\sin {\phi_j}& \cos{\phi_j}
\end{array}\right]\left[\begin{array}{*{2}{c}}
\cos {\theta_j}& \sin {\phi_j}\\
-\sin{\theta_j}& \cos{\phi_j}
\end{array}\right]= \left[\begin{array}{*{2}{c}}
1& \sin{(\theta_j-\phi_j)}\\
\sin{(\theta_j-\phi_j)}& 1
\end{array}\right]$
\end{center}

\noindent
\\
Obviously, the value of activation operator depends on the values of $\theta_j$  and $\phi_j$, which in turn depend on the initial weight operators and the training data.

\section {Conclusion and Future Work}
\noindent
This paper presented a novel algorithm achieves autonomous quantum perceptron neural network (AQPNN) to enable real time computations. The developed algorithm represents a good computational alternate to the classical perceptron neural network approach. AQPNN constructs self-adaptive activation operators capable to accomplish the learning process in limited number of iterations and reduces the overall computational cost. These activation operators can be applied in both quantum and classical applications. The efficiency of the proposed algorithm is evaluated via solving four different problems where promising and comparable results are given. In addition, to train the AQPNN algorithm, it uses a limited number of training data samples, and in testing, it accomplishes a well generalization. Using the proposed perceptron in real world applications is one of our future aims.

\newpage

\begin{appendices}
\setcounter{equation}{0}
\renewcommand\theequation{\thesection\arabic{equation}}
\section{}
\noindent
\emph{\textbf{Qubit:}}
The smallest element store information in quantum computer is called quantum-bit (qubit). The qubit takes either value of 0 or 1 or a superposition of these states in the form :
\begin{equation}
{\ket{\psi}}= a \ket{0}+ b \ket{1}. \label{qubits}
\end{equation}
Where a,b are complex numbers called the probability amplitudes. The qubit state ${\ket{\psi}}$ is collapse into either basis state $\ket{0}$ or $\ket{1}$ with probability ${\left|a\right|}^{2}$ or ${\left|b\right|}^{2}$  respectively where
\begin{equation}
{\left|a\right|}^{2}+{\left|b\right|}^{2}=1. \label{quprob}
\end{equation}
\noindent
\\
\emph{\textbf{Hadamard product (matrices)}}
In mathematics, the Hadamard product is a binary operation that takes two matrices of the same dimensions, and produces another matrix where each element i,j is the product of elements i,j of the original two matrices. For two matrices, A, B of the same dimension,
m x n the Hadamard product, $ A \circ B$, is a matrix, of the same dimension as the operands, with elements given by
\begin{equation}
(A \circ B)_{i,j}=(A)_{i,j} \dot (B)_{i,j}
\,. \label{qubitsss}
\end{equation}

\emph{Example} : Suppose two matrices \begin{center}
$A=\left[\begin{array}{*{2}{c}}
1& 2\\
3& 4
\end{array}\right]$, and
$B=\left[\begin{array}{*{2}{c}}
5& 6\\
7& 8
\end{array}\right]$
\end{center}
then the hadamrd product is
\begin{center}
$A \circ B= \left[\begin{array}{*{2}{c}}
1(5)& 2(6)\\
3(7)& 4(8)
\end{array}\right]=
\left[\begin{array}{*{2}{c}}
5& 12\\
21& 32
\end{array}\right]$
\end{center}

\end{appendices}

\end{document}